\documentclass[10.5pt, conference, compsocconf]{IEEEtran}
\ifCLASSINFOpdf
\else
\fi
\usepackage{hyperref}
\usepackage{graphicx}
\usepackage{amsmath}
\usepackage{dirtytalk}
\graphicspath{Figures}

\begin{document}
%
\title{A BLSTM  Network for Printed Bengali OCR System with High Accuracy}


\author{
\IEEEauthorblockA{Debabrata Paul\\
Society for Natural Language Technology Research\\
Kolkata, India\\
debabratapaul90@gmail.com}
\and
\IEEEauthorblockA{Bidyut B. Chaudhuri\\
Techno India University\\
Kolkata, India\\
bbcisical@gmail.com}
}


%


\maketitle

\begin{abstract}
This paper presents a printed Bengali and English text OCR system developed by us using a single hidden BLSTM-CTC architecture having 128 units. Here, we did not use any peephole connection and dropout in the BLSTM, which helped us in getting better accuracy. This architecture was trained by 47,720 text lines that include English words also. When tested over 20 different Bengali fonts, it has produced character level accuracy of 99.32\% and word level accuracy of 96.65\%. A good Indic multi script OCR system is also developed by Google. It sometimes recognizes a character of Bengali into the same character of a non-Bengali script, especially Assamese, which has no distinction from Bengali, except for a few characters. For example, Bengali character for `RA' is sometimes recognized as that of Assamese, mainly in conjunct consonant forms. Our OCR is free from such errors. This OCR system is available online at
\url{https://banglaocr.nltr.org}

\end{abstract}


%
\IEEEpeerreviewmaketitle

\section{Introduction}
OCR is among the oldest and most successful pattern recognition and machine learning based application technology, which is being improved till today. Early approaches for printed texts OCR of document image were done by statistical and SVM classifiers, among others. Most of them worked after line, word and character segmentation, which incurred a fair amount of mistakes at these stages, more so on Alpha-syllabary or Abugida scripts like Devanagari and Bengali, among others. Those pre-processing errors used to substantially reduce the overall OCR accuracy. Later on, word to character segmentation was avoided by Hidden Markov Model (HMM) based OCRs, which used hidden states with prior and transition probabilities during classifier design. However, to the best of our knowledge, a full text line (containing 10-12 words) based OCR with HMM system alone that work with very high accuracy was not reported in the literature. More recently, Drastic improvement in the performance was noted after the use of Recurrent Neural Net (RNN) based classifier. Among RNNs, Long Short-Term Memory (LSTM) architecture, originally proposed by Hochreiter and Schmidhuber~\cite{Hochreiter1997}, is a multi-gated Recurrent Neural Net (RNN) which has better ability of exploiting long range contexts and can be made deep structured. Schuster and Paliwal~\cite{650093} have made the LSTM architecture more powerful by adding bi-directionality (which utilize past and future context) in their problem of Speech processing. This network is now called Bidirectional LSTM or BLSTM, which along with Connectionist Temporal Classification (CTC) layer~\cite{Graves:2006:CTC:1143844.1143891} have been successfully employed in both printed and online/offline handwritten text OCR in various scripts~\cite{Graves:2008:OHR:2981780.2981848},~\cite{NIPS2007_3213}.

Most of the early works on printed OCR system were done on English and related Alphabetic writing systems as well as Chinese, Japanese and Korean scripts. The works on Indian scripts were somewhat lagging behind and in 1998, the first workable system on Indic script (Bengali) OCR was reported~\cite{ChaudhuriP98}, which was followed by similar studies on other scripts during the new millennium~\cite{DBLP:conf/sspr/LehalS02},~\cite{906135},~\cite{1227699}. All these methods worked in the character level, where word and character segmentation from the document text lines were required.

Reports on full page printed script OCR, using Neural net architecture started about ten years ago. Breuel et al.~\cite{6628705} have proposed one LSTM based system to recognize English and Fraktur printed scripts. It is a open-source software, to be available online~\cite{ocropus}. For Thai printed script recognition, LSTM with vertical component shifting for compound characters has been reported recently~\cite{Emsawas:2016:TPC:3118016.3118026}. Moreover, Adnan Ul-Hasan et al.~\cite{6628777} have proposed a Bi-directional LSTM network based printed Urdu Nastaleeq script recognition. Among Indian scripts, Shankaran and Jawahar~\cite{6460137} was perhaps the first to report a printed Devanagari word OCR system using a BLSTM Network (which was provided to them by Alex Graves). Later on, this group extended the work for some other Indic scripts as well~\cite{6830986}. Ray et al.~\cite{7050699} also used Deep BLSTM Network for recognizing printed Oriya script words. Mathew et al.~\cite{7490115} have proposed LSTM based multilingual word based Indic OCR. All these Indic script based OCRs were word-based systems, where text lines and words were to be identified from the document image. So, for a document page line and word isolation errors used to contribute in the overall recognition error. Later on, a text line-based approach for printed Devanagari script recognition was proposed by Karayil et al.~\cite{7333901}. Recently, Biswas et al.~\cite{Biswas} have used a hybrid architecture for recognition of printed text lines in degraded documents. Chavan et al.~\cite{8313738} have shown line based text recognition for Indian scripts including Bengali, but their accuracy was relatively poor. There is a open-source multi-script system (including Bengali) called Tesseract~\cite{tesseract} (now owned by Google), but its accuracy is not so high for Bengali script.

Most of the above studies are done at the laboratory level, which were not tested extensively by a large number of users. But apart from Tesseract, Google developed another good OCR system in 2016 which can be used for documents having text in multiple Indic scripts, including Devanagari and Bengali. To use it freely, the scanned document image file needs to be submitted online into the `Google drive' and the OCR output is obtained as paragraph-wise UNICODE file which can be printed in any UNICODE supported Bengali font. For fair quality Indian alpha-syllabary printed scripts including Bengali, the Google OCR shows very low character and word level error in normal document images. It is also quite robust for noisy and degraded documents. But, a person cannot employ the Google system in a stand-alone mode or integrate it as a `sub-system' into some other bigger application system, since it is not sold as a software unit. Also, there are some other limitations of using this system, which are described in section IV. These facts  motivated us to develop an indigenous Bengali (with English) OCR system, whose initial system and its performance are reported here.

Our OCR system is also based on BLSTM, where English texts are included because many Bengali documents contain English words printed in English script. Also, our OCR is a line based system, so in the pre-processing stage, only individual lines are to be separated from the given text image. The CTC has powerful capability to automatically align ground truth of a line with the text line image, so that the system learns from lines and do not need words and characters segmentation for the inputs.



\section{Printed Bengali Scripts}

Machine Printing of Bengali script dates back to 1778 AD, when a book named \say{A grammar of the Bengali language} authored by N. B. Halhed was published at Hoogly. For this book Charles Wilkins created Bengali letterforms with the help of native smith Panchanan Karmakar. Later on, Serampore Missionaries Press played a major role in improving Bengali fonts while printing Christian religious books in Bengali. Uniform and good quality typefaces were generated and good quality Bengali printing started from the end to first quarter of nineteenth century. Further improvements in characters forms were done by great educationist and social reformer Iswarchandra Vidyasagar. In his honour, the typefaces generated since 1850s are now called as Vidyasagari fonts. Some typefaces of compound components of this form become obsolute after spelling correction movement of Bengali words in twentieth century. In 1935, the publishers of popular Bengali newspaper Anandabazar Patrika (ABP) planned to use the Linotype machine for composing Bengali letters. For that, they had to make major changes in Bengali typefaces. Thus, the Linotype Bengali font came in 1950s and books, magazines and newspapers were published in that font. Later on, in another machine called Monotype, a similar font was also made for Bengali composing. We may call them together as Lino-Monotype fonts. Some publishers used such font for about 20 years. But the majority of printing houses and publishers continued to use Vidyasagari fonts with small change influenced by lino-Monotype font. From 1980s computer based system started to be used for Bengali printing and the use of Lino-Monotype fonts almost stopped. Because of their short span of use, we did not include Linotype font based prints for training our OCR network.

\section{Proposed method}
At first, Depending on the document image quality, preprocessing like noise reduction, skew correction etc. are done by the off-the-shelf methods. As stated before, our method is a text-line based OCR system. So, we need to identify individual text lines from the document image. To do this, some classical text line detection methods such as horizontal projection histogram or Hough transform and their variations exist in the literature. Among more recent approaches, Rashid et al.~\cite{6195344} developed a Multi-Layer Perceptron (MLP) and Hidden Markov Model (HMM) based text line separation method. Also, Zhi Tian et al.~\cite{ctpn} have proposed a better CNN based network named Connectionist Text Proposal Network (CTPN) for line detection from scene text. It works very well for English scripts. We tested this CTPN, trained for English text line detection, directly on Bengali document images. It was working well, but not to the extent of our previous unpublished method developed by a non-NN approach which is briefly described below.

Our method works in two phases. In the first phase, the document vertical strip based projection profile valleys of document sections were considered as initial guess for the line end/beginning. In the second phase, this guess was refined whereby object image and other artifacts were discarded. Two or more touching text lines and extra thin line generated wrongly in the first stage were also corrected at this stage.

\begin{figure}[h]
\centering
  \includegraphics[width=3.28 in]{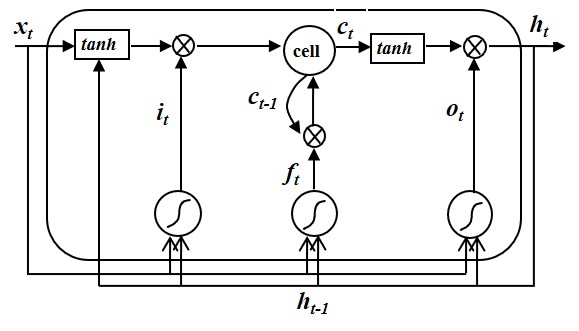}
  \caption{A folded Long Short Term Memory Block
  (without peephole connection)}
  \label{fig:Fig1}
\end{figure}

Once the individual text lines are obtained, each line image is normalized to a standard height of 48 pixels. Our experience on Bengali OCR research for the last 20 years is that normalized line height of 48 pixels contains enough information for text recognition. Next, the height normalized line image is presented to the BLSTM-CTC based OCR system for training and validation. After this, it is used for testing. For training, validation and testing however, we generated the line-wise ground truths in Bengali Unicode. The ground-truth of test lines are actually used for performance evaluation. We generated the BLSTM network using Tensorflow Toolkit (Version 1.3). This network is primarily inspired by the model of~\cite{6707742}.
The basic unit of LSTM used here is shown in Figure 1. In general, an LSTM unit has 3 peephole connections from its cell to input, forget and output gates. But we did not use any peephole connection because in our case, deletion of peephole reduced the CTC loss substantially.

\begin{figure}[h]
\centering
  \includegraphics{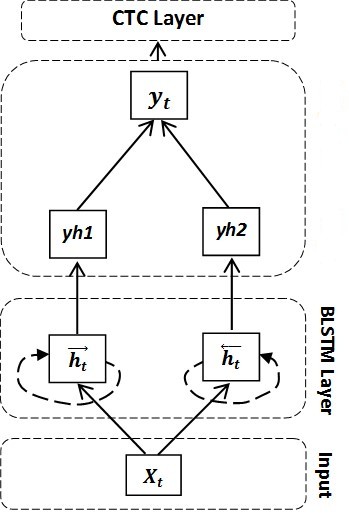}
  \caption{Our network model.}
  \label{fig:Fig2}
\end{figure}

Since we have used Bidirectional system, there are two LSTM sub-layers, namely forward and backward sub-layers. At present, each sub-layer contains 128 LSTM units. The outputs from the units of corresponding sub-layers are combined through weights and biases. These weights and biases were initialized by Xavier Initializer~\cite{pmlr-v9-glorot10a}, which lead to fast convergence during training phase. This layer is followed by the Connectionist Temporal Classification (CTC) layer, which is the output layer. In this layer, the number of units is equal to the number of class labels and one extra unit for the blank label. The output of CTC is like a probability distribution over all label sequence including blank label for each frame of the input sequences. During training, the objective function of CTC calculates the loss, which is the negative log probability of target labels. During training, the target is to minimize the loss by altering the weights of network by back-propagation. We used momentum optimizer with Learning rate 0.0001 and Momentum 0.9 for back-propagation. During testing, the CTC acts as the classifier. It produces the most promising labels for a given input sequence as the final output. We use CTC Beam search decoder for getting the output. Figure 2 demonstrates our overall network architecture.

As stated before, for the sequence labeling task, the Bi-directional LSTM contains two separate hidden layers. In Figure 2 the `dashed' circular lines with arrow shaped end in BLSTM layer indicate the rollover of LSTM units. If unrolled, the units can be perceived as shown in Figure 3. It processes the input sequence in both forward and backward directions. The two hidden layers are connected to a single output layer, thereby providing access to both past and future contexts. Thus, including the input and output layer, our system has three basic layers only. We have chosen the number of units in each sub-layer of the BLSTM layer as 128 and only one bi-directional hidden layer. One of the reasons of this is lack of high-end hardware availability.

In the absence of peephole connections like Figure 1, for a single LSTM unit, the input output relations are described by the following equations-
\begin{align*}
i_{t} &=\sigma(W_{i x} x_{t}+W_{i h} h_{t-1}+b_{i})\\[1.7mm]
f_{t} &=\sigma(W_{f x} x_{t}+W_{f h} h_{t-1}+b_{f})\\[1.7mm]
c_{t} &=f_{t} \odot c_{t-1}+i_{t} \odot \tanh (W_{c x} x_{t}+W_{c h} h_{t-1}+b_{c}) \\[1.7mm]
o_{t} &=\sigma(W_{o x} x_{t}+W_{o h} h_{t-1}+b_{o}) \\[1.7mm]
h_{t} &=o_{\mathrm{t}} \odot \tanh (c_{t})
\end{align*}
where the \(W\) terms denote elements of connection weight matrices (e.g. \(W_{i x}\) is the weight matrix from the input gate to the input), the b terms denote bias vectors (e.g. \(b_{i}\) is the input gate bias vector), \(h\) is the output from the unit, \(\sigma\) is the logistic sigmoid function, and \(i\), \(f\), \(o\) and \(c\) are the input gate, forget gate, output gate and memory cell vectors, respectively and \(\odot\) is the element-wise multiplication between vectors.

\begin{figure}
\centering
  \includegraphics{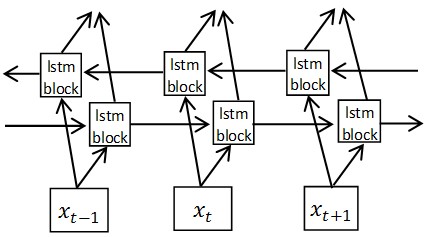}
  \caption{Sequence of Bidirectional LSTM.}
  \label{fig:Fig3}
\end{figure}
For an input sequence \(x = (x_{1}, x_{2},\dots, x_{t})\), the hidden sequence \(h = (h_{1}, h_{2},\dots, h_{t})\) and the output sequence \(y = (y_{1}, y_{2},\dots, y_{t})\) the following relation holds. Here \(\overrightarrow{\boldsymbol{h}_{t}}\) corresponds to the forward LSTM sequence and \(\overleftarrow{\boldsymbol{h}_{t}}\) corresponds to the backward LSTM sequence. The final output \(y_{t}\) is obtained by adding weighted sum of both along with bias \(b_{y}\), as given in the equations below.

\begin{align*}
\overrightarrow{\boldsymbol{h}_{t}} &=f(W_{x \overrightarrow{h}} x_{t}+W_{\overrightarrow{h} \overrightarrow{h}} \overrightarrow{h}_{t-1}+b_{\overrightarrow{h}})\\[1.7mm]
\overleftarrow{\boldsymbol{h}_{t}} &=f(W_{x \overleftarrow{h}} x_{t}+W_{\overleftarrow{h} \overleftarrow{h}} \overleftarrow{h}_{t-1}+b_{\overleftarrow{h}})\\[1.7mm]
yh1 &=W_{\overrightarrow{h} y} \overrightarrow{h_{t}}\\[1.7mm]
yh2 &=W_{\overleftarrow{h} y} \overleftarrow{h_{t}}\\[1.7mm]
y_{t} &=y h 1+y h 2+b_{y}\\
\end{align*}

\section{Experiment and results}
Our network model is learned with \(47,720\) text line images (which includes \(4,500\) lines containing English). Also, \(1,500\) line images are used for validation and \(6,645\) line images are used for testing. These lines may contain mixture of Bengali and English words. These are equivalent to \(4,72,167\) words or \(28,67,659\) characters of training set, \(14,365\) words or \(87,607\) characters of validation set and \(61,582\) words or \(3,69,931\) characters of test set. The text images are collected from various categories of books, letters, newspapers, notices, magazines etc. All images are scanned at 300 dpi (dots per inch). We have included only commonly used fonts (about 20 in number) in our training set. Some training samples are shown in Figure 4. For training, we used UTF-8 format to prepare the ground-truth files. Each ground-truth file corresponds to one training line of the document image. We did not alter Unicode ordering while preparing the ground-truth.

\begin{figure}[!ht]
\centering
  \includegraphics[width=3.28 in]{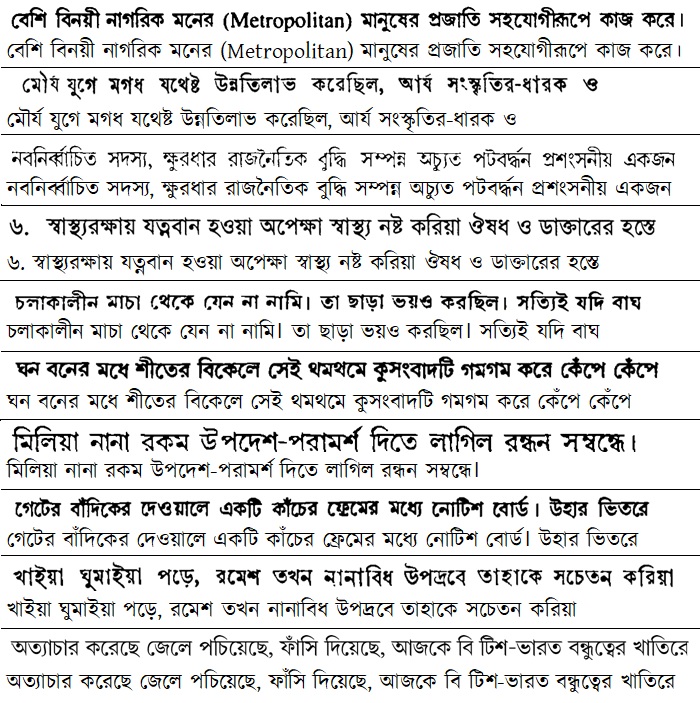}
  \caption{Training samples. Image lines (odd numbered) and corresponding ground truths (even numbered), expressed in single Bengali font.}
  \label{fig:Fig4}
\end{figure}

A total of \(166\) class labels (165 + 1 for CTC blank label) are used for classification. The class labels correspond to all Bengali vowels and consonants, commonly used symbols and punctuation marks, uppercase and lowercase English characters in UNICODE.

Some readers may be aware that Bengali is an alpha-syllabary script consisting of vowels, consonants, vowel markers, compound characters and numerals consisting of more than 300 character shapes. These, along with English characters, punctuation marks and numerals need about 400 character shapes to be recognized (which includes `danri' or `danda', the sign that represents full-stop in Bengali and some other scripts). So, some readers may wonder how these 400 shapes can be accommodated by only 166 output labels. Actually, the trick lies in the UNICODE rule of representing about 230 Bengali compound characters. A compound character in UNICODE is the combination of UNICODEs of two, three and four consonants joined by the code of a sign called `hasant' or `halant' between every two consecutive consonants. When these are output from CTC, these trigger the generation of the appropriate compound character in a UNICODE compliant Bengali font. 

As stated before, training and test image height of a line is normalized to 48 pixel. For training, we provide one strip of size 1 $\times$ 48 pixels gray values each time as input feature vector to the BLSTM. Our batch size for the training is one. One training epoch is completed when all input lines are processed. We preset the maximum number of training epochs as 80. An upper bound to 80 epochs was chosen experimentally to be sure that we are not stuck in a local error minimum. At each epoch, we compute the CTC loss and validation error. The validation error is calculated by summing over CTC loss on 1500 validation samples. The readers should not confuse these with text recognition error rate. The training is stopped when the difference in both CTC loss and validation error between epoch \(t\) and \((t+1)\) are less than or equal to 0.01 and 0.1, respectively.

We have tested our BLSTM Layer model with 80, 100, 128 and 156 hidden units. Among these, 128 hidden units at 48-th epoch produced minimum error on the validation set. So, we chose 128 units trained at 48-th epoch for the purpose of testing. We expected better results in case of 156 hidden units. Failure to this may be due to fairly small size of training samples, since training of BLSTM-CTC architecture needs huge amount of data, which we could not afford to produce. Epoch-wise CTC loss and validation error are represented graphically in Figure 5 and Figure 6, respectively.

\begin{figure}
\centering
  \includegraphics[width=3.28 in]{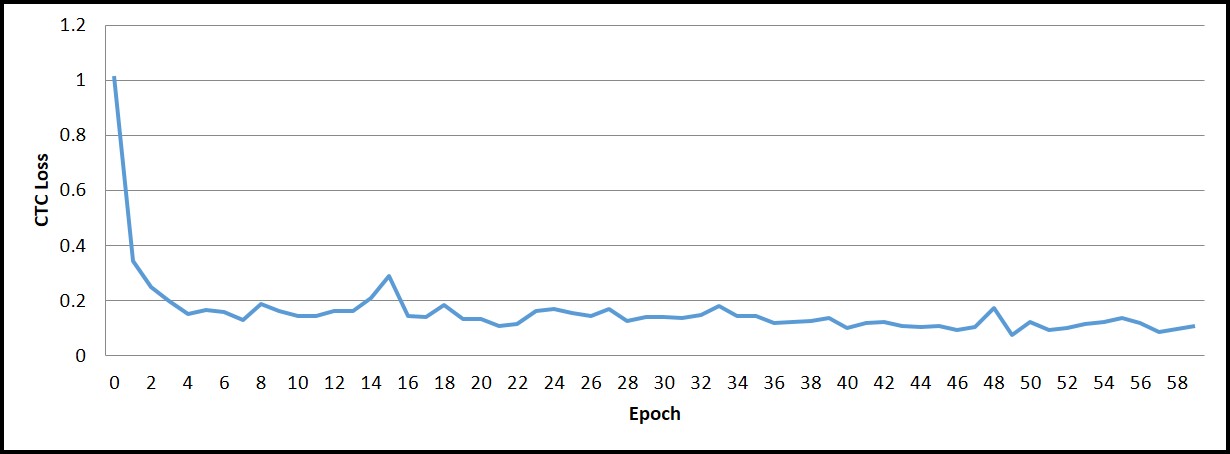}  
  \caption{Epoch Vs CTC Loss for 128 units.}
  \label{fig:Fig5}
\end{figure}

\begin{figure}
\centering
  \includegraphics[width=3.28 in]{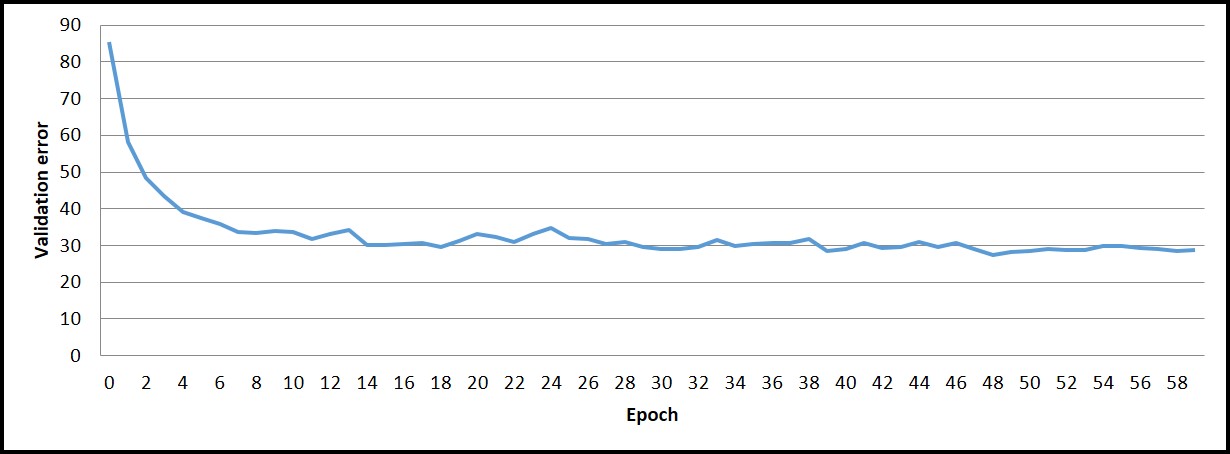}
  \caption{Epoch Vs Validation error for 128 units.}
  \label{fig:Fig6}
\end{figure}

We evaluated our model on the test set, which is calculated as percentage of Character-level Accuracy (CA) and Word-level Accuracy (WA) obtained by Minimum Edit Distance (MED) method. Thus, the percent of CA and WA are calculated by the following relations.

\begin{align*}
CA &=(1-\frac{\text {Sum of all character level MEDs}}{\text {Total number of characters}}) * 100\\ \\
WA &=(1-\frac{\text {Sum of all word level MEDs}}{\text {Total number of words}}) * 100
\end{align*}

The test set was also run on tesseract 4.0 LSTM version and Google drive OCR system. The experimental results are shows in Table I.

\begin{table}[!ht]
\centering
\caption{Test Results}
\begin{tabular}{|l|c|c|ll}
\cline{1-3}
\multicolumn{1}{|c|}{System/method} & Character Accuracy \% & Word Accuracy \% &  &  \\ \cline{1-3}
Tesseract 4.0                       & 91.79                 & 76.31            &  &  \\ \cline{1-3}
Google drive                        & 98.54                 & 92.86            &  &  \\ \cline{1-3}
Our method                          & 99.32                 & 96.65            &  &  \\ \cline{1-3}
\end{tabular}
\end{table}

We have compared our work with two systems, both owned by Google, since we did not find any other open software in the net. One of the Google system is Tesseract 4.0, which has LSTM engine and is freely available. We call the other and more accurate system that can be run through the Google drive as Google Drive-based System (GDS). It works in the paragraph level of the document text. As mentioned earlier, this system API cannot be used in standalone mode or cannot be used like a sub-system in a bigger application. Our experimental results are shown in Table I, where the results of our system is somewhat better than GDS system. However, the GDS is more versatile since it can OCR on a single document containing multiple major Indian scripts and English. This advantage has a flip side also. There are some character shapes for two or more scripts which are extremely similar looking. For example, except for two basic characters, the shapes of all other basic characters look exactly same in Bengali and Assamese, but their Unicode are different. Thus, if precaution is not taken, the chance of Unicode of a character in Bengali being output as the Unicode of the same character in Assamese is fairly high. The GDS suffers to some extent by this problem since a single GDS system works on both Bengali and Assamese text. Thus, the visual results in Google may be much better than what the Unicode based MED results show. Also, Google drive OCR sometimes create unnecessary Danda. This problem was not noticed in our system. Since our system does not include Devanagari OCR.

\section{Conclusion}
A simplified one BLSTM layer based printed Bengali (with English) OCR system has been developed with high character and word level accuracy. Also, the training set of data required here is reasonably low. Our future work is to use multi-layer (deep) architecture trained with bigger database and to make the system more versatile with respect to document quality. Moreover, we shall be working towards combining our classical method with CTPN for more accurate text line identification from more complex documents. We also plan to include Assamese script in our system. Moreover, we intend to enhance our system to recognize the Bengali text printed in obsolete Lino-Monotype fonts.
Our fully operational system is available online at\\ \textbf{\url{https://banglaocr.nltr.org}}






%




\bibliographystyle{IEEEtran}
\bibliography{reference}

\begin{thebibliography}{10}
\providecommand{\url}[1]{#1}
\csname url@samestyle\endcsname
\providecommand{\newblock}{\relax}
\providecommand{\bibinfo}[2]{#2}
\providecommand{\BIBentrySTDinterwordspacing}{\spaceskip=0pt\relax}
\providecommand{\BIBentryALTinterwordstretchfactor}{4}
\providecommand{\BIBentryALTinterwordspacing}{\spaceskip=\fontdimen2\font plus
\BIBentryALTinterwordstretchfactor\fontdimen3\font minus
  \fontdimen4\font\relax}
\providecommand{\BIBforeignlanguage}[2]{{%
\expandafter\ifx\csname l@#1\endcsname\relax
\typeout{** WARNING: IEEEtran.bst: No hyphenation pattern has been}%
\typeout{** loaded for the language `#1'. Using the pattern for}%
\typeout{** the default language instead.}%
\else
\language=\csname l@#1\endcsname
\fi
#2}}
\providecommand{\BIBdecl}{\relax}
\BIBdecl

\bibitem{Hochreiter1997}
\BIBentryALTinterwordspacing
S.~Hochreiter and J.~Schmidhuber, ``Long short-term memory,'' \emph{Neural
  Comput.}, vol.~9, no.~8, pp. 1735--1780, Nov. 1997. [Online]. Available:
  \url{http://dx.doi.org/10.1162/neco.1997.9.8.1735}
\BIBentrySTDinterwordspacing

\bibitem{650093}
M.~{Schuster} and K.~K. {Paliwal}, ``Bidirectional recurrent neural networks,''
  \emph{IEEE Transactions on Signal Processing}, vol.~45, no.~11, pp.
  2673--2681, Nov 1997.

\bibitem{Graves:2006:CTC:1143844.1143891}
\BIBentryALTinterwordspacing
A.~Graves, S.~Fern\'{a}ndez, F.~Gomez, and J.~Schmidhuber, ``Connectionist
  temporal classification: Labelling unsegmented sequence data with recurrent
  neural networks,'' in \emph{Proceedings of the 23rd International Conference
  on Machine Learning}, ser. ICML '06.\hskip 1em plus 0.5em minus 0.4em\relax
  New York, NY, USA: ACM, 2006, pp. 369--376. [Online]. Available:
  \url{http://doi.acm.org/10.1145/1143844.1143891}
\BIBentrySTDinterwordspacing

\bibitem{Graves:2008:OHR:2981780.2981848}
\BIBentryALTinterwordspacing
A.~Graves and J.~Schmidhuber, ``Offline handwriting recognition with
  multidimensional recurrent neural networks,'' in \emph{Proceedings of the
  21st International Conference on Neural Information Processing Systems}, ser.
  NIPS'08.\hskip 1em plus 0.5em minus 0.4em\relax USA: Curran Associates Inc.,
  2008, pp. 545--552. [Online]. Available:
  \url{http://dl.acm.org/citation.cfm?id=2981780.2981848}
\BIBentrySTDinterwordspacing

\bibitem{NIPS2007_3213}
A.~Graves, M.~Liwicki, H.~Bunke, J.~Schmidhuber, and S.~Fern\'{a}ndez,
  ``Unconstrained on-line handwriting recognition with recurrent neural
  networks,'' in \emph{Advances in Neural Information Processing Systems 20},
  J.~C. Platt, D.~Koller, Y.~Singer, and S.~T. Roweis, Eds.\hskip 1em plus
  0.5em minus 0.4em\relax Curran Associates, Inc., 2008, pp. 577--584.

\bibitem{ChaudhuriP98}
\BIBentryALTinterwordspacing
B.~B. Chaudhuri and U.~Pal, ``A complete printed bangla {OCR} system,''
  \emph{Pattern Recognition}, vol.~31, no.~5, pp. 531--549, 1998. [Online].
  Available: \url{https://doi.org/10.1016/S0031-3203(97)00078-2}
\BIBentrySTDinterwordspacing

\bibitem{DBLP:conf/sspr/LehalS02}
\BIBentryALTinterwordspacing
G.~S. Lehal and C.~Singh, ``A complete {OCR} system for gurmukhi script,'' in
  \emph{Structural, Syntactic, and Statistical Pattern Recognition, Joint
  {IAPR} International Workshops {SSPR} 2002 and {SPR} 2002, Windsor, Ontario,
  Canada, August 6-9, 2002, Proceedings}, 2002, pp. 358--367. [Online].
  Available: \url{https://doi.org/10.1007/3-540-70659-3\_37}
\BIBentrySTDinterwordspacing

\bibitem{906135}
G.~S. {Lehal} and C.~{Singh}, ``A gurmukhi script recognition system,'' in
  \emph{Proceedings 15th International Conference on Pattern Recognition.
  ICPR-2000}, vol.~2, Sep. 2000, pp. 557--560 vol.2.

\bibitem{1227699}
C.~V. {Jawahar}, M.~N. S. S.~K. {Pavan Kumar}, and S.~S. {Ravi Kiran}, ``A
  bilingual ocr for hindi-telugu documents and its applications,'' in
  \emph{Seventh International Conference on Document Analysis and Recognition,
  2003. Proceedings.}, Aug 2003, pp. 408--412 vol.1.

\bibitem{6628705}
T.~M. {Breuel}, A.~{Ul-Hasan}, M.~A. {Al-Azawi}, and F.~{Shafait},
  ``High-performance ocr for printed english and fraktur using lstm networks,''
  in \emph{2013 12th International Conference on Document Analysis and
  Recognition}, Aug 2013, pp. 683--687.

\bibitem{ocropus}
\BIBentryALTinterwordspacing
``Ocropus - open source document analysis and ocr system.'' [Online].
  Available: \url{https://code.google.com/p/ocropus}
\BIBentrySTDinterwordspacing

\bibitem{Emsawas:2016:TPC:3118016.3118026}
T.~Emsawas and B.~Kijsirikul, ``Thai printed character recognition using long
  short-term memory and vertical component shifting,'' in \emph{Proceedings of
  the 14th Pacific Rim International Conference on Trends in Artificial
  Intelligence}, ser. PRICAI'16.\hskip 1em plus 0.5em minus 0.4em\relax
  Switzerland: Springer, 2016, pp. 106--115.

\bibitem{6628777}
A.~{Ul-Hasan}, S.~B. {Ahmed}, F.~{Rashid}, F.~{Shafait}, and T.~M. {Breuel},
  ``Offline printed urdu nastaleeq script recognition with bidirectional lstm
  networks,'' in \emph{2013 12th International Conference on Document Analysis
  and Recognition}, Aug 2013, pp. 1061--1065.

\bibitem{6460137}
N.~{Sankaran} and C.~V. {Jawahar}, ``Recognition of printed devanagari text
  using blstm neural network,'' in \emph{Proceedings of the 21st International
  Conference on Pattern Recognition (ICPR2012)}, Nov 2012, pp. 322--325.

\bibitem{6830986}
P.~{Krishnan}, N.~{Sankaran}, A.~K. {Singh}, and C.~V. {Jawahar}, ``Towards a
  robust ocr system for indic scripts,'' in \emph{2014 11th IAPR International
  Workshop on Document Analysis Systems}, April 2014, pp. 141--145.

\bibitem{7050699}
A.~{Ray}, S.~{Rajeswar}, and S.~{Chaudhury}, ``Text recognition using deep
  blstm networks,'' in \emph{2015 Eighth International Conference on Advances
  in Pattern Recognition (ICAPR)}, Jan 2015, pp. 1--6.

\bibitem{7490115}
M.~{Mathew}, A.~K. {Singh}, and C.~V. {Jawahar}, ``Multilingual ocr for indic
  scripts,'' in \emph{2016 12th IAPR Workshop on Document Analysis Systems
  (DAS)}, April 2016, pp. 186--191.

\bibitem{7333901}
T.~{Karayil}, A.~{Ul-Hasan}, and T.~M. {Breuel}, ``A segmentation-free approach
  for printed devanagari script recognition,'' in \emph{2015 13th International
  Conference on Document Analysis and Recognition (ICDAR)}, Aug 2015, pp.
  946--950.

\bibitem{Biswas}
C.~Biswas, P.~Sarathi~Mukherjee, K.~Ghosh, U.~Bhattacharya, and S.~K.~Parui,
  ``A hybrid deep architecture for robust recognition of text lines of degraded
  printed documents,'' 08 2018, pp. 3174--3179.

\bibitem{8313738}
V.~{Chavan}, A.~{Malage}, K.~{Mehrotra}, and M.~K. {Gupta}, ``Printed text
  recognition using blstm and mdlstm for indian languages,'' in \emph{2017
  Fourth International Conference on Image Information Processing (ICIIP)}, Dec
  2017, pp. 1--6.

\bibitem{tesseract}
\BIBentryALTinterwordspacing
``Tesseract - tesseract open source ocr engine.'' [Online]. Available:
  \url{https://github.com/tesseract-ocr/tesseract}
\BIBentrySTDinterwordspacing

\bibitem{6195344}
S.~F. {Rashid}, F.~{Shafait}, and T.~M. {Breuel}, ``Scanning neural network for
  text line recognition,'' in \emph{2012 10th IAPR International Workshop on
  Document Analysis Systems}, March 2012, pp. 105--109.

\bibitem{ctpn}
Z.~Tian, W.~Huang, H.~Tong, P.~He, and Y.~Qiao, ``Detecting text in natural
  image with connectionist text proposal network,'' vol. 9912, 10 2016, pp.
  56--72.

\bibitem{6707742}
A.~{Graves}, N.~{Jaitly}, and A.~{Mohamed}, ``Hybrid speech recognition with
  deep bidirectional lstm,'' in \emph{2013 IEEE Workshop on Automatic Speech
  Recognition and Understanding}, Dec 2013, pp. 273--278.

\bibitem{pmlr-v9-glorot10a}
\BIBentryALTinterwordspacing
X.~Glorot and Y.~Bengio, ``Understanding the difficulty of training deep
  feedforward neural networks,'' in \emph{Proceedings of the Thirteenth
  International Conference on Artificial Intelligence and Statistics}, ser.
  Proceedings of Machine Learning Research, Y.~W. Teh and M.~Titterington,
  Eds., vol.~9.\hskip 1em plus 0.5em minus 0.4em\relax Chia Laguna Resort,
  Sardinia, Italy: PMLR, 13--15 May 2010, pp. 249--256. [Online]. Available:
  \url{http://proceedings.mlr.press/v9/glorot10a.html}
\BIBentrySTDinterwordspacing

\end{thebibliography}

\end{document}